\documentclass[a4paper,twoside]{article}

\usepackage{epsfig}
\usepackage[nolist]{acronym}
\usepackage{caption}
\usepackage{subcaption}
\usepackage{calc}
\usepackage{amssymb}
\usepackage{amstext}
\usepackage{amsmath}
\usepackage{amsthm}
\usepackage{color}
\usepackage{xcolor}
\usepackage{pifont}
\usepackage{hyperref}
\usepackage{pslatex}
\usepackage{apalike}
\usepackage{algorithm2e}
\usepackage[bottom]{footmisc}
\usepackage{todonotes}
\usepackage{booktabs}
\usepackage{multicol}
\usepackage{multirow}
\usepackage{amssymb}
\newcommand{\cmark}{\color{green!40!black}{\ding{51}}}
\newcommand{\xmark}{\color{red}{\ding{55}}}
\usepackage{SCITEPRESS}    
\usepackage{cellspace}
\newcommand\cincludegraphics[2][]{\raisebox{-0.3\height}{\includegraphics[#1]{#2}}}

\begin{acronym}
    \acro{CNN}{convolutional neural network} 
\end{acronym}

\tikzstyle{ioblock} = [rectangle, rounded corners, text centered, draw =black!80]
\tikzstyle{operation} = [circle, text centered, draw =black!80]
\tikzstyle{block} = [rectangle, draw =black!80, text width=1.5cm, text centered]
\tikzstyle{convolution} = [rectangle, rounded corners, draw=black!80, fill=blue!20, text width=1.5cm, text centered]
\tikzstyle{tip} = [draw, -latex]

\begin{document}

\title{When Medical Imaging Met Self-Attention: A Love Story That Didn't Quite Work Out}

\author{\authorname{Tristan Piater\sup{1}\orcidAuthor{0009-0008-1938-6261}, Niklas Penzel\sup{1}\orcidAuthor{0000-0001-8002-4130}, Gideon Stein\sup{1}\orcidAuthor{0000-0002-2735-1842} and Joachim Denzler \sup{1}\orcidAuthor{0000-0002-3193-3300}}
\affiliation{\sup{1}Computer Vision Group, Friedrich Schiller University, Jena, Germany}
\email{\{tristan.piater, niklas.penzel, gideon.stein, joachim.denzler\}@uni-jena.de}
}


\abstract{A substantial body of research has focused on developing systems that assist medical professionals during labor-intensive early screening processes, many based on convolutional deep-learning architectures.
Recently, multiple studies explored the application of so-called self-attention mechanisms in the vision domain.
These studies often report empirical improvements over fully convolutional approaches on various datasets and tasks.
To evaluate this trend for medical imaging, we extend two widely adopted convolutional architectures with different self-attention variants on two different medical datasets.
With this, we aim to specifically evaluate the possible advantages of additional self-attention.
We compare our models with similarly sized convolutional and attention-based baselines and evaluate performance gains statistically.
Additionally, we investigate how including such layers changes the features learned by these models during the training.
Following a hyperparameter search, and contrary to our expectations, we observe no significant improvement in balanced accuracy over fully convolutional models. 
We also find that important features, such as dermoscopic structures in skin lesion images, are still not learned by employing self-attention.
Finally, analyzing local explanations, we confirm biased feature usage. 
We conclude that merely incorporating attention is insufficient to surpass the performance of existing fully convolutional methods.}

\onecolumn \maketitle \normalsize \setcounter{footnote}{0} \vfill

\section{\uppercase{Introduction}}
\label{sec:introduction}

Computer vision models can aid medical practitioners by providing visual analysis of skin lesions or tumor tissues.
In both cases, a correct classification can help save lives since the survival rate of cancer patients increases drastically when their condition is detected early.
Malignant melanomata, for example, still result in over 7000 deaths in the US per year \cite{skinCan}.
Hence, a substantial body of research, e.g., \cite{Tschandl2018-fg,mishra2016overview,codella2019skin,celebi2019dermoscopy,BreastCancer,Bera2019ArtificialII,generalizedFramework}, has been focused on developing automated systems that examine medical images to assist medical practitioners with the demanding screening process.
Usually, those models are built on convolutional neural networks (CNNs), a well-established deep-learning archetype.

Previous work investigated \ac{CNN} models and their feature usage \cite{reimers2021conditional,penzel2022investigating} specifically for skin lesion classification and found that these networks learn only a subset of medically relevant features.
To be specific, of the dermatological ABCD rule \cite{nachbar1994abcd}, only \textbf{A}symmetry and \textbf{B}order irregularity were incorporated consistently in the prediction process of state-of-the-art melanoma models.
Further, \textbf{C}olor and \textbf{D}ermoscopic structures were often not learned.
Additionally, modern automatic skin lesion classifiers often overfit on biases contained in the training data~\cite{reimers2021conditional}, e.g., spurious colorful patches \cite{scope2016study}.

Recently, emerging from natural language processing \cite{vaswani2017attention}, multiple studies explored the application of so-called self-attention mechanisms in the vision domain \cite{kolesnikov:vit,Wang18:NLNN,Ram19:SASA}.
These studies often report empirical improvements over state-of-the-art approaches on various datasets and tasks.
Following the intuition that self-attention mechanisms possibly allow for a better comprehension of global features,  e.g., dermoscopic structures in skin lesions or number of cells in tumor tissue, previous approaches directly evaluated popular transformer architectures on medical image classification tasks \cite{Yang23:vitSC,HE2022102357,krishna23:lesionAid}.
While they typically report an improvement in accuracy, it is not self-evident that the observed improvement is due to the attention mechanisms and not other circumstances, e.g., possibly increased parameter counts or altered input representation \cite{DBLP:journals/corr/abs-2201-09792,DBLP:journals/corr/abs-2105-01601}
which often go hand in hand when deploying transformer architectures. 
Furthermore, we find that the question of whether self-attention mechanisms can help to learn additional medically relevant features is understudied.
To answer these questions, we study the possible benefits of self-attention mechanisms on two medical classification tasks: skin lesion classification \cite{web:isic} and tumor tissue classification \cite{bandi2018:detection}.
Specifically, we extend two widely adopted \ac{CNN}-backbones: ResNet \cite{He15:resNet} and EfficientNet \cite{Tan19:effNet} with different self-attention variants and evaluate in-distribution and out-of-distribution performance. Further, we compare our models with convolutional and vision transformer baselines.
In the literature, two strategies have been proposed for incorporating self-attention into \acp{CNN}. 
The global self-attention approach \cite{Wang18:NLNN} and the local self-attention block \cite{Ram19:SASA}, which we adopt and extend. 
Furthermore, we also investigate combinations of these approaches.
For all our architectures, we perform hyperparameter searches and report the parameter count to enable fair comparisons.

Contrary to our expectations, we found that the tested self-attention mechanisms do not significantly improve the performance of medical image classifiers, nor do vision transformers when kept at a similar parameter count as the convolutional baselines.
In contrast, we sometimes measure a significant decrease in performance.

With these results in mind, we derive global and local explanations of our models' behavior. 
To be specific, for the skin lesion models using the global explainability method described in \cite{reimers2020determining}.
Here, we directly follow the approaches described in \cite{reimers2021conditional,penzel2022investigating}.
We find that adding self-attention does generally not help in learning medical-relevant features.
While sometimes additional medically relevant features are learned, this is often accompanied by relying on more biases.

Furthermore, we derive local explanations for a selection of images using both Grad-CAM \cite{selvaraju2017grad} as well as visualizing the learned global attention maps where applicable.
While self-attention natively provides local explanations that can lead to the discovery of structural biases, we find that out-of-the-box methods, we used Grad-CAM \cite{selvaraju2017grad}, can provide very similar insights.
Also, observing these insights, we find that the global self-attention maps in our experiments often reveal unwanted attention spikes and artifacts in the background.
Similar behavior was also observed for other transformer models, e.g., in \cite{darcet2023vision,xiao2023efficient}.

To conclude: our empirical analysis suggests that merely incorporating attention is insufficient to surpass the performance of existing fully convolutional methods. 

\section{\uppercase{Approach}}
\label{sec:method}

To explain our setup, we first describe both, global~\cite{Wang18:NLNN} and local~\cite{Ram19:SASA}, approaches to extend \acp{CNN} with self-attention methods.
Further, we propose another local self-attention implementation for a smoother integration into pre-trained networks.
Afterward, we detail how we include those attention methods into two selected \acp{CNN}: ResNet18 \cite{He15:resNet} and EfficientNet-B0 \cite{Tan19:effNet}.

\subsection{\uppercase{Global Self-Attention:}}
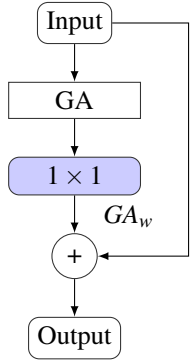
\begin{figure}
   \begin{center} 
        \begin{tikzpicture}[node distance = 0.5cm,auto]
            \node[ioblock] (input) {Input};
            \node[block] (ga) [below=of input] {GA};
            \node[convolution] (weights) [below=of ga, label=below right:$GA_w$] {\(1 \times 1\)};
            \node[operation] (add) [below=of weights] {+};
            \node[ioblock] (out) [below=of add] {Output};
            
            \path[tip] (input) -- (ga);
            \path[tip] (ga) -- (weights);
            \path[tip] (weights) -- (add);
            \draw[tip] (input.east) |- (1.5,0) |- (add.east);
            \path[tip] (add) -- (out);
        \end{tikzpicture}
    \end{center}
    \caption{Computational graph of the global self-attention}
    \label{gsa_impl}
\end{figure}

We implement the global self-attention (GA) method following the details in~\cite{Wang18:NLNN}.
To add further details, it is a global operation where each of the output values depends on all of the input vectors.
Therefore, it should increase the capability of capturing long-range dependencies, which are especially helpful in the medical area.
An example here would be global features like skin lesion asymmetry or dermoscopic structures.
Behind the GA operation, a zero-initialized $1\times 1 \times 1$ $GA_w$ convolution is added.
This convolution, together with a residual connection, results in an identity mapping at initialization \cite{Wang18:NLNN}.
Hence, inserting it into a network keeps the pre-trained behavior but somewhat increases the parameter count.
The computational graph can be seen in Figure~\ref{gsa_impl}.

\subsection{\uppercase{Local Self-Attention:}}
Secondly, we investigate a local self-attention (LA) mechanism based on~\cite{Ram19:SASA}.
This version does not operate on the whole feature map for each input pixel, but rather on the local neighborhood \(N_k(i,j) = \{a,b \in \mathbb{Z}: |a - i | < \frac{k}{2} \land |b - j| < \frac{k}{2}\}\), where $k$ determines the neighborhood size.
The authors introduce this approach by replacing the final convolutional layers of a \ac{CNN}.
The implementation is based on three $1 \times 1$ convolutions, which require fewer parameters compared to a $k \times k (k > 1)$ convolution.
Another advantage is its content-based computation.
However, it is not capable of imitating the computation of the convolutional layer it replaces.
Therefore, adding it to a \ac{CNN} destroys the pre-trained behavior.

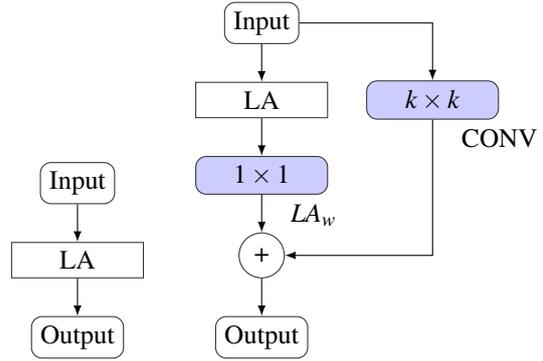
\begin{figure}
    \begin{subfigure}{0.13\textwidth}
       \begin{center} 
            \begin{tikzpicture}[node distance = 0.5cm,auto]
                \node[ioblock] (input) {Input};
                \node[block] (la) [below=of input] {LA};
                \node[ioblock] (out) [below=of la] {Output};
                
                \path[tip] (input) -- (la);
                \path[tip] (la) -- (out);
            \end{tikzpicture}
        \end{center}
        \caption{LA}
        \label{sfig:la}
    \end{subfigure}
    \begin{subfigure}{0.35\textwidth}
       \begin{center} 
            \begin{tikzpicture}[node distance = 0.5cm,auto]
                \node[ioblock] (input) {Input};
                \node[block] (la) [below=of input] {LA};
                \node[convolution] (conv) [right=of la, label=below right:CONV] {\(k \times k\)};
                \node[convolution] (weights) [below=of la, label=below right:$LA_w$] {\(1 \times 1\)};
                \node[operation] (add) [below=of weights] {+};
                \node[ioblock] (out) [below=of add] {Output};
                
                \path[tip] (input) -- (la);
                \path[tip] (la) -- (weights);
                \path[tip] (weights) -- (add);
                \draw[tip] (input.east) -| (conv.north);
                \draw[tip] (conv.south) |- (add.east);
                \path[tip] (add) -- (out);
            \end{tikzpicture}
        \end{center}
        \caption{ELA}
        \label{sfig:ela}
    \end{subfigure}
    \caption{Computational graphs for both local approaches: local self-attention (LA) and embedded local self-attention (ELA).
    }
    \label{ela_impl}
\end{figure}

To overcome this issue, we propose an alternative implementation, which we call embedded local self-attention, henceforth ``ELA''.
The implementation is inspired by the design of GA described above.
Specifically, we add a residual connection around the introduced LA block, containing the original convolutions and pre-trained weights.
Similarly to GA, we add a zero-initialized $1\times 1\times 1$ convolution after the LA block to conserve pre-trained behavior.
Hence, it is content-based and keeps pre-trained behavior at initialization, at the cost of adding four $1 \times 1$ convolutions.
Both operations are visualized in Figure~\ref{ela_impl} as computational graphs, showing the similarity of ELA to GA.

\subsection{\uppercase{Implementation Details}}
\label{sec:impl-details}

We base our analysis on two commonly used \ac{CNN} architectures, ResNet~\cite{He15:resNet} and EfficientNet~\cite{Tan19:effNet}.
For all models, we rely on commonly used ImageNet \cite{russakovsky2015imagenet} pre-trained weights as initialization.
In the following, we describe our specific architecture choices. 

\subsubsection{\uppercase{ResNet} \cite{He15:resNet}}
Regarding ResNet architectures, we opt for ResNet18, a well-established and frequently used variant.
We insert the GA block after the first and second blocks of ResNet18, as recommended in~\cite{Wang18:NLNN}.
Regarding local attention, for both LA and ELA, we replace all convolutional layers of the last block.
A similar setup is discussed in \cite{Ram19:SASA}.
Note that all ResNet architectures feature four blocks.
Hence, our inclusion of attention mechanisms could be implemented similarly for other variants.

\subsubsection{\uppercase{EfficientNet} \cite{Tan19:effNet}}
Again, we choose the smallest variant of this model family, EfficientNet-B0, featuring seven blocks. 
To insert the GA component, we again follow the recommendations of~\cite{Wang18:NLNN} and include it after the second and third architecture blocks.
Similarly, we add LA and ELA by replacing convolutions in the last block.
Note that EfficientNet-B0 utilizes so-called MBConv blocks \cite{sandler2018mobilenetv2}.
Hence, minimizing the number of parameters and increasing efficiency. 
An MBConv block contains a convolutional layer using groups, amongst other techniques, to reduce parameters.
Hence, naively replacing this grouped convolution significantly increases the parameter count.
To avoid this issue, we replace the whole MBConv block instead, leading to a similar parameter reduction as observed for the ResNet models described above.
We argue that this parameter reduction aligns better with the reasoning stated in \cite{Ram19:SASA} and increases comparability between our selected \ac{CNN} architectures.

\subsubsection{\uppercase{Model Selection}}
\label{sec:extendedModels}

\begin{table}
    \centering
    \caption{Parameter counts of all architectures (in million) and relative change compared to the corresponding baselines.}
    \label{tab:para_count}
    \setlength{\tabcolsep}{4pt}
    \begin{tabular}{l|cc}
        & ResNet18 & EfficientNet-B0\\
        \hline
        Base & $11.18${\tiny $\pm 0.00\%$} & $4.01${\tiny $\pm 0.00\%$}  \\
        + GA & $11.22${\tiny $+0.37\%$} & $4.02${\tiny $+0.12\%$} \\
        + LA & $5.67${\tiny $-49.25\%$} & $3.48${\tiny $-13.29\%$} \\
        + GA + LA & $5.71${\tiny $-48.87\%$} & $3.48${\tiny $-13.17\%$} \\
        + ELA & $14.98${\tiny $+34.02\%$} & $4.30${\tiny $+7.16\%$} \\
        + GA + ELA & $15.02${\tiny $+34.40\%$} & $4.30${\tiny $+7.27\%$} \\
        \hline
        ViT & \multicolumn{2}{c}{$5.49${\tiny $\pm 0.00\%$}}\\
    \end{tabular}
\end{table}

Table~\ref{tab:para_count} summarizes the different architectures we investigate in this study to analyze the influence of self-attention on skin lesion and tissue classification.
Additionally, we report the parameter counts to ensure that none of our architecture changes widely skew the model capacities.
For comparison, we also list our selected baselines: ResNet18 \cite{He15:resNet}, EfficientNet-B0 \cite{Tan19:effNet}, and two attention-only ViT models \cite{kolesnikov:vit}.
Here, we select the ViT tiny variant with a parameter count of 5.49 million parameters which is in between both selected \ac{CNN} architectures.
We investigate a version pre-trained on the commonly used ImageNet 1K challenge \cite{russakovsky2015imagenet} to stay comparable to our other experiments and baselines, as well as one ViT pre-trained on the larger ImageNet 21K dataset. 
We add the latter model, given empirical observations that larger pre-training datasets increase the downstream performance of transformer models \cite{kolesnikov:vit}.

\section{\uppercase{Experiments}}
\label{sec:exp}
We start by describing the general experimental setup we employed before discussing the achieved results.
We then analyze the feature usage of the best-performing skin lesion models.
Finally, we visualize the learned attention maps and compare them to local Grad-CAM \cite{selvaraju2017grad} explanations to gather additional insights. 

\subsection{\uppercase{Experimental Setup}}
\label{sec:exp_setup}

To investigate the influence of additional self-attention blocks, we analyze the performance of our hybrid approaches in three distinct ways.
First, we trained the models on skin-lesion and tumor tissue classification, comparing the balanced accuracies on in-distribution and out-of-distribution test sets.
Concerning skin lesions, we use data from the ISIC archive~\cite{web:isic} as training data and rely on the PAD-UFES-20 dataset~\cite{PACHECO2020106221} as out-of-distribution (OOD) test data.
To align these tasks, we simplified the classification problem to the three classes benign nevi, melanoma, and others.
Regarding tumor tissue classification, we use the Camelyon17 dataset~\cite{bandi2018:detection}, which already provides an out-of-distribution split in addition to the training data.
The domain shift in Camelyon17 is introduced by sampling the OOD set from a different hospital.

To increase reliability, we performed a small hyperparameter search, considering the learning rate and weight decay. 
To reduce the search space, we fixed our batch size to 32. 
For data augmentation, we followed the recommendations from~\cite{Perez18:dataAug} and chose their best-performing augmentation scheme.
This scheme randomly applies cropping, affine transformations, flipping, and changing the hue, brightness, contrast, and saturation of the input.
In our experiments, we use SGD as an optimizer.
To determine the performance of a specific hyperparameter combination, we perform the search with 2 random data splits.

After determining the optimal learning rate and weight decay for each model, which is shown in Table~\ref{tab:hypers}, we train each architecture 10 times on different data splits to determine the mean and standard deviation of the achieved performances. 
Additionally, we test for statistical significance by performing Welch's T-Test~\cite{Welch47:ttest} on these samples.
Here, we employ the widely used significance level of $p<0.05$.

\begin{table*}[t]
    \centering
    \caption{Balanced Accuracies on each Dataset and each Model. Values inside the parentheses show the balanced accuracies on the OOD-dataset. Bold values indicate \textbf{statistically significant} differences to the respective base model.}
    \begin{tabular}{l|cc|cc}
        & \multicolumn{2}{c|}{ISIC Dataset} & \multicolumn{2}{c}{Camelyon17}\\
        & ResNet & EfficientNet & ResNet & EfficientNet \\
        \hline
        Base & $73.9${\tiny $\pm 1.74$} ($57.4${\tiny $\pm 8.87$})&$75.4${\tiny $\pm 1.75$} ($57.6${\tiny $\pm 15.18$})&$98.4${\tiny $\pm 0.18$} ($94.4${\tiny $\pm 1.66$}) & $98.5${\tiny $\pm 0.16$} ($94.5${\tiny $\pm 2.33$}) \\
        +GA & $\mathbf{72.1}${\tiny $\pm 1.31$} ($56.3${\tiny $\pm 12.92$})&$76.6${\tiny $\pm 1.83$} ($55.8${\tiny $\pm 12.67$})&$98.4${\tiny $\pm 0.07$} ($94.0${\tiny $\pm 1.73$}) & $98.6${\tiny $\pm 0.17$} ($95.4${\tiny $\pm 0.24$}) \\
        +LA & $\mathbf{70.8}${\tiny $\pm 2.96$} ($56.3${\tiny $\pm 5.83$})&$75.5${\tiny $\pm 1.58$} ($55.5${\tiny $\pm 7.77$})&$98.2${\tiny $\pm 0.36$} ($\mathbf{92.9}${\tiny $\pm 3.15$}) & $98.4${\tiny $\pm 0.21$} ($94.3${\tiny $\pm 2.01$}) \\
        +GA+LA & $\mathbf{71.2}${\tiny $\pm 1.44$} ($55.5${\tiny $\pm 4.64$})&$75.8${\tiny $\pm 1.14$} ($55.4${\tiny $\pm 12.48$})&$98.4${\tiny $\pm 0.16$} ($\mathbf{92.2}${\tiny $\pm 5.90$}) & $98.6${\tiny $\pm 0.26$} ($94.3${\tiny $\pm 0.63$}) \\
        +ELA & $\mathbf{72.3} ${\tiny $\pm 1.45$} ($58.4${\tiny $\pm 7.05$})&$73.8${\tiny $\pm 2.29$} ($57.9${\tiny $\pm 8.43$})&$98.2${\tiny $\pm 0.23$} ($\mathbf{91.4}${\tiny $\pm 8.22$}) & $\mathbf{98.3}${\tiny $\pm 0.07$} ($94.1${\tiny $\pm 2.09$}) \\
        +GA+ELA & $73.5${\tiny $\pm 1.35$} ($58.8${\tiny $\pm 10.41$})&$75.9${\tiny $\pm 1.77$} ($54.7${\tiny $\pm 11.84$})&$98.4${\tiny $\pm 0.14$} ($95.1${\tiny $\pm 0.44$}) & $98.5${\tiny $\pm 0.40$} ($94.3${\tiny $\pm 3.01$}) \\
        \hline
        \hline
        & 1k & 21k & 1k & 21 k\\
        ViT & $66.0${\tiny $\pm 17.36$} ($52.6${\tiny $\pm 113.45$}) & $75.0${\tiny $\pm 0.90$} ($56.7${\tiny $\pm 6.94$}) & $98.3${\tiny $\pm 0.43$} ($94.8${\tiny $\pm 0.91$}) & $98.3${\tiny $\pm 0.11$} ($94.5${\tiny $\pm 0.96$}) \\
    \end{tabular}
    \label{tab:empiciral}
\end{table*}
\subsection{\uppercase{Performance Analysis}}

Table~\ref{tab:empiciral} summarizes the balanced accuracy for all examined models and corresponding baselines on both datasets and the corresponding OOD sets.

On the ISIC archive data, the ResNet18 variants with added attention mechanisms generally underperformed compared to the convolutional baseline.
Further, while the EfficientNet-B0 variants primarily exhibit improvements, on average, over the baseline, these improvements are not statistically significant.
On the Camelyon17 dataset, we observed similar performances between the baselines and our hybrid models, with the only significant result being the EfficientNet + ELA, which performs worse on average.
Hence, overall, we found no reliable upward trend in performance when including self-attention mechanisms in a model. 

We, however, noticed smaller differences between the specific attention versions.
Specifically, we found that local approaches show, on average, increased performance in combination with GA.
Nevertheless, the overall performance still decreases compared to the baselines.

While the significant reduction of performance for some models could be attributed to the reduced parameter count (LA models), we also observe similar behavior in models with increased parameter count (ELA).
This is further corroborated by the observed higher performance of the EfficientNet models in the skin lesion task despite a lower parameter count compared to the ResNet variants.

Regarding OOD performance, we see that there are substantial differences between the datasets.
For the Camelyon17 dataset, the degradation of performance is minimal for all models, and they still achieve high balanced accuracies.
We observe higher average OOD performance for the EfficientNet + GA hybrid model. 
However, this increase is not statistically significant.
Further, the only statistically reliable results regarding the OOD performance are the observed decreases for some of the ResNet hybrid models.
In the skin lesion classification task, the drop in performance is much more severe, also leading to higher standard deviations.
Hence, we did not observe any significant improvements of the self-attention hybrid models over the fully convolutional baselines in this task.

The performance of the ViT heavily depends on the pretraining data size in the skin lesion task.
While pretraining, the model with the 21k classes version of ImageNet~\cite{russakovsky2015imagenet} outperforms the ResNet baseline in the skin lesion classification task, pretraining with default ImageNet 1K does perform significantly worse. 
This result confirms previous observations, e.g., \cite{kolesnikov:vit}, that larger pretraining datasets can increase downstream performance.
In our experiment, this specifically meant that two of our ten ViT 1K models diverged, resulting in random guessing capabilities and explaining the huge observed standard deviation.

\subsection{\uppercase{Global Explanations - Feature Usage}}
\label{sec:exp_features}

To further investigate the model adaptations described in Sec.~\ref{sec:extendedModels}, we investigated our hybrid models and baselines on a feature level in the skin lesion classification task.
Here we employ the feature attribution method described in \cite{reimers2020determining}.
Previous work \cite{reimers2021conditional,penzel2022investigating} also applied this method to skin lesion classification and investigated the usage of features related to the ABCD rule \cite{nachbar1994abcd} and known biases in skin lesion data \cite{mishra2016overview}.

Here, we specifically follow \cite{reimers2021conditional,penzel2022investigating} and extract the same four features related to the ABCD rule, namely \textbf{A}symmetry, \textbf{B}order irregularity, \textbf{C}olor, and \textbf{D}ermoscopic structures, as well as the four bias features age of a patient, sex of the patient, skin color, and the occurrence of large colorful patches \cite{scope2016study}.
For a detailed description of these features, we refer the reader to the original work \cite{reimers2021conditional}.
Furthermore, we follow the hyperparameter settings described in \cite{penzel2022investigating} and select three conditional independence tests, namely cHSIC \cite{fukumizu2007kernel}, RCoT \cite{strobl2019approximate}, and CMIknn \cite{runge2018conditional}.
We report the majority decision and use a significance level of $p<0.01$ \cite{penzel2022investigating}.

\begin{table*}
    \centering
    \setlength{\tabcolsep}{8pt}
    \caption{
    Feature usage of our best-performing skin lesion models according to balanced accuracy. 
    We abbreviate \textbf{A}symmetry, \textbf{B}order irregularity, \textbf{C}olor, and \textbf{D}ermoscopic structures with the associated ABCD rule letter.
    We use significance level $p < 0.01$ \cite{reimers2021conditional}. 
    }
    \label{tab:bal-features}
    \begin{tabular}{cl>{\centering\arraybackslash\hspace{0pt}}p{0.8cm}>{\centering\arraybackslash\hspace{0pt}}p{0.8cm}>{\centering\arraybackslash\hspace{0pt}}p{0.8cm}>{\centering\arraybackslash\hspace{0pt}}p{0.8cm}>{\centering\arraybackslash\hspace{0pt}}p{0.8cm}>{\centering\arraybackslash\hspace{0pt}}p{0.8cm}cc}
        \toprule
         & & \multirow{2}{*}{\textbf{A}} & \multirow{2}{*}{\textbf{B}} & \multirow{2}{*}{\textbf{C}} & \multirow{2}{*}{\textbf{D}} & \multirow{2}{*}{Age} & \multirow{2}{*}{Sex} & Skin & Colorful\\
         & Model &  &  &  &  & &  & color & patches \\
        \midrule
        \multirow{6}{*}{\rotatebox[origin=c]{90}{ResNet}} 
        & Base & \cmark & \cmark & \xmark & \xmark & \cmark & \cmark & \cmark & \cmark \\
        & +GA & \cmark & \cmark & \cmark & \xmark & \cmark & \cmark & \cmark & \cmark \\
        & +LA & \xmark & \cmark & \xmark & \xmark & \cmark & \cmark & \cmark & \cmark \\
        & +GA+LA & \cmark & \cmark & \cmark & \xmark & \cmark & \xmark & \cmark & \cmark \\
        & +ELA & \cmark & \cmark & \xmark & \xmark & \cmark & \cmark & \cmark & \cmark \\
        & +GA+ELA & \cmark & \cmark & \xmark & \xmark & \cmark & \cmark & \cmark & \cmark \\
         \midrule
        \multirow{6}{*}{\rotatebox[origin=c]{90}{EfficientNet}} 
        & Base & \cmark & \cmark & \xmark & \xmark & \cmark & \xmark & \cmark & \xmark \\
        & +GA & \xmark & \cmark & \xmark & \xmark & \cmark & \xmark & \cmark & \cmark \\
        & +LA & \cmark & \cmark & \xmark & \xmark & \cmark & \cmark & \cmark & \cmark \\
        & +GA+LA & \cmark & \cmark & \cmark & \xmark & \cmark & \xmark & \cmark & \cmark \\
        & +ELA & \cmark & \cmark & \xmark & \xmark & \cmark & \cmark & \cmark & \xmark \\
        & +GA+ELA & \xmark & \cmark & \xmark & \xmark & \cmark & \cmark & \cmark & \cmark \\
         \midrule
         \\[-0.9em]
        \multirow{2}{*}{\rotatebox[origin=c]{90}{ViT}}
         & 1K & \cmark & \cmark & \cmark & \xmark & \cmark & \xmark & \cmark & \cmark \\
         \\[-0.9em]
         & 21K & \cmark & \cmark & \cmark & \xmark & \cmark & \cmark & \cmark & \cmark \\
         \\[-0.9em]
        \bottomrule
    \end{tabular}
\end{table*}

Table~\ref{tab:bal-features} contains the results of our feature analysis.
Regarding the bias features, the age and skin color of the patients are both learned by all analyzed models.
The colorful patches and patient's sex features are also often learned, however, predominantly by the ResNet variants.
The EfficientNet models seem to be more robust in that regard.
Looking at the meaningful ABCD rule features, we observe that the models often learn the asymmetry or border irregularity while the color feature is rarely, and the dermoscopic structures feature is never learned.
In contrast, both ViT models learn to incorporate the color of the lesion into their decision process.
In general, GA seems to produce slightly better results, given that all extended models that utilize the color feature also include GA.
However, this observation is inconsistent as other combinations of GA do not result in this behavior.
Some models containing GA regress and do not learn to utilize the skin lesion asymmetry, which is consistently learned by the \ac{CNN} baselines and which has also been reported previously \cite{reimers2021conditional}.

To conclude, we do not find a systematic improvement of the extended models incorporating attention mechanisms over the fully convolutional baselines.
There is a small benefit in some attention-based models with respect to the color feature, especially for the ViT architecture.
Nevertheless, these results are similar to the empirical evaluation described above, where some of the extended models achieved a performance increase, which was not statistically significant.

\begin{table*}
    \centering    
    \caption{Qualitative examples of explanations generated using either Grad-CAM \cite{selvaraju2017grad} or by visualizing the learned attention maps. We choose one image per class per dataset and the best performing models of either the baselines or adapted with GA or ELA (here abbreviated with G and E).}
    \label{tab:att-vis}
    \begin{tabular}{ccc|Sc Sc Sc Sc| Sc Sc Sc}
        & & & \multicolumn{4}{c}{ISIC} &  \multicolumn{3}{|c}{Camelyon17}\\
        \multicolumn{3}{c}{Model} & Melanoma & Nevus & Other & Mean & Tumor & No Tumor & Mean \\
        \midrule\\
        \multirow[c]{8}{*}[-3.45cm]{\rotatebox[origin=c]{90}{Grad-CAM}} 
        & \multirow[c]{4}{*}[-1.3cm]{\rotatebox[origin=c]{90}{ResNet}}
            & \rotatebox[origin=c]{90}{-} & \cincludegraphics[width=.08\textwidth]{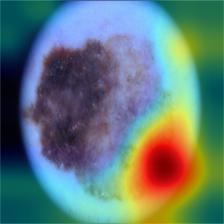} &\cincludegraphics[width=.08\textwidth]{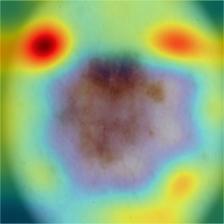}&\cincludegraphics[width=.08\textwidth]{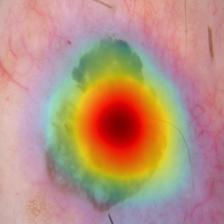}&\cincludegraphics[width=.08\textwidth]{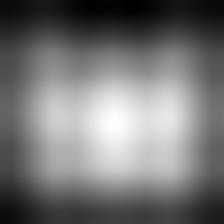}&\cincludegraphics[width=.08\textwidth]{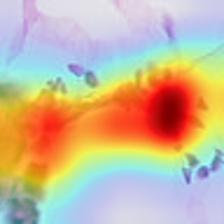}&\cincludegraphics[width=.08\textwidth]{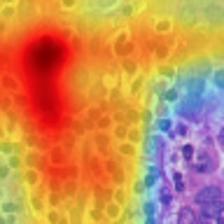}&\cincludegraphics[width=.08\textwidth]{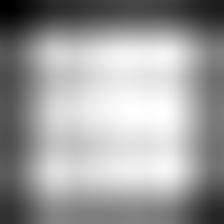}\\
            & & \rotatebox[origin=c]{90}{+G} & \cincludegraphics[width=.08\textwidth]{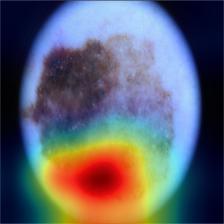} &\cincludegraphics[width=.08\textwidth]{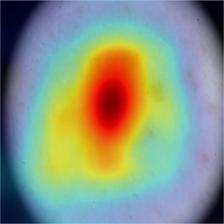}&\cincludegraphics[width=.08\textwidth]{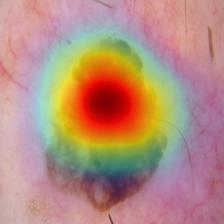}&\cincludegraphics[width=.08\textwidth]{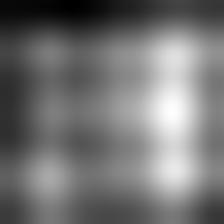}&\cincludegraphics[width=.08\textwidth]{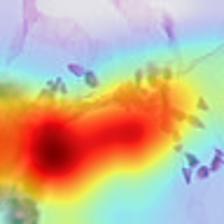}&\cincludegraphics[width=.08\textwidth]{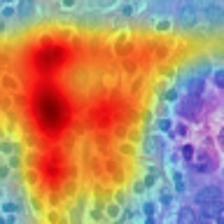}&\cincludegraphics[width=.08\textwidth]{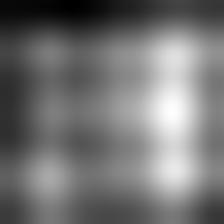}\\
            & & \rotatebox[origin=c]{90}{+E} & \cincludegraphics[width=.08\textwidth]{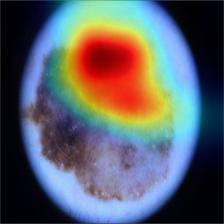} &\cincludegraphics[width=.08\textwidth]{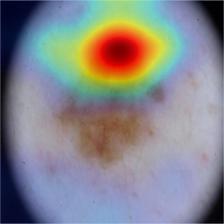}&\cincludegraphics[width=.08\textwidth]{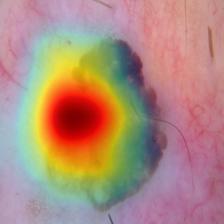}&\cincludegraphics[width=.08\textwidth]{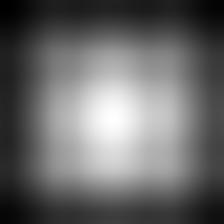}&\cincludegraphics[width=.08\textwidth]{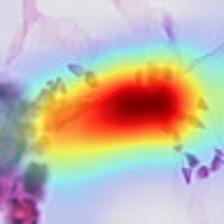}&\cincludegraphics[width=.08\textwidth]{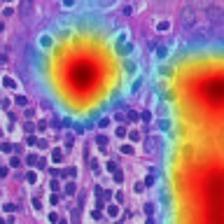}&\cincludegraphics[width=.08\textwidth]{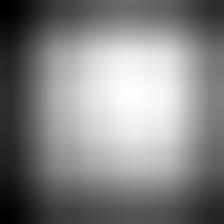}\\
            & & \rotatebox[origin=c]{90}{+GE} & \cincludegraphics[width=.08\textwidth]{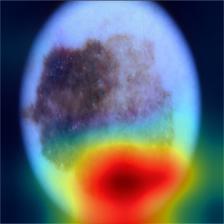} &\cincludegraphics[width=.08\textwidth]{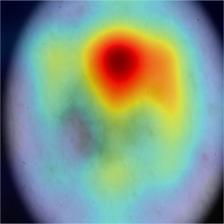}&\cincludegraphics[width=.08\textwidth]{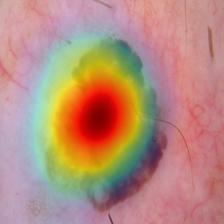}&\cincludegraphics[width=.08\textwidth]{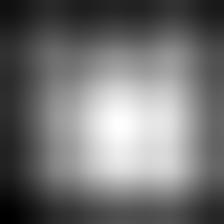}&\cincludegraphics[width=.08\textwidth]{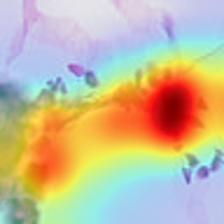}&\cincludegraphics[width=.08\textwidth]{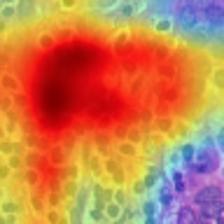}&\cincludegraphics[width=.08\textwidth]{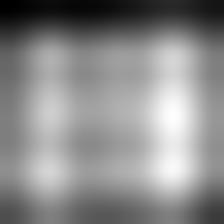}\\
            \cmidrule{2-10}
        & \multirow{4}{*}[-1.3cm]{\rotatebox[origin=c]{90}{EfficientNet}}
            & \rotatebox[origin=c]{90}{-} & \cincludegraphics[width=.08\textwidth]{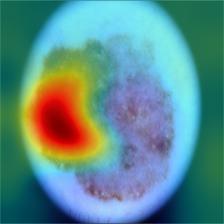} &\cincludegraphics[width=.08\textwidth]{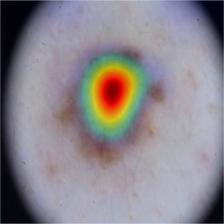}&\cincludegraphics[width=.08\textwidth]{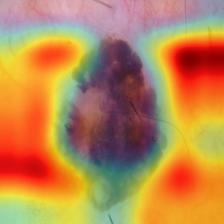}&\cincludegraphics[width=.08\textwidth]{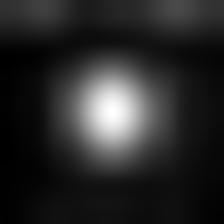}&\cincludegraphics[width=.08\textwidth]{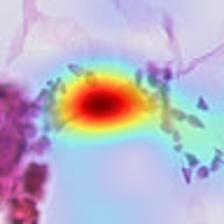}&\cincludegraphics[width=.08\textwidth]{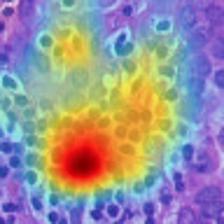}&\cincludegraphics[width=.08\textwidth]{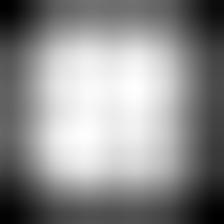}\\
            & & \rotatebox[origin=c]{90}{+G} & \cincludegraphics[width=.08\textwidth]{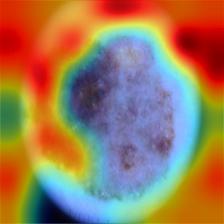} &\cincludegraphics[width=.08\textwidth]{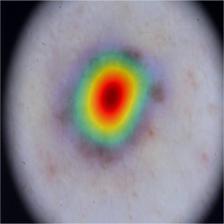}&\cincludegraphics[width=.08\textwidth]{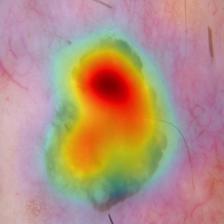}&\cincludegraphics[width=.08\textwidth]{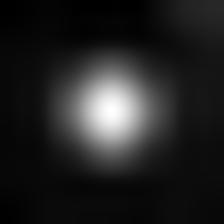}&\cincludegraphics[width=.08\textwidth]{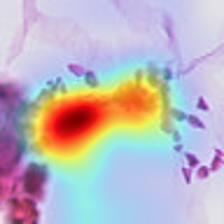}&\cincludegraphics[width=.08\textwidth]{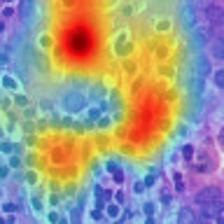}&\cincludegraphics[width=.08\textwidth]{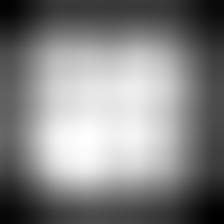}\\
            & & \rotatebox[origin=c]{90}{+E} & \cincludegraphics[width=.08\textwidth]{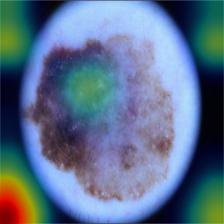} &\cincludegraphics[width=.08\textwidth]{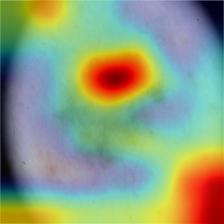}&\cincludegraphics[width=.08\textwidth]{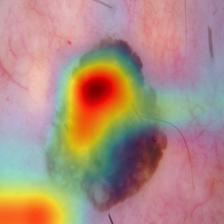}&\cincludegraphics[width=.08\textwidth]{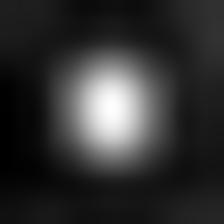}&\cincludegraphics[width=.08\textwidth]{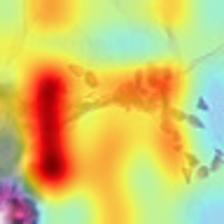}&\cincludegraphics[width=.08\textwidth]{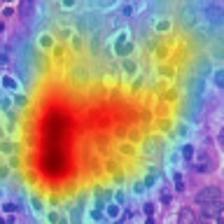}&\cincludegraphics[width=.08\textwidth]{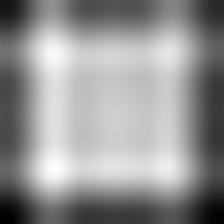}\\
            & & \rotatebox[origin=c]{90}{+GE} & \cincludegraphics[width=.08\textwidth]{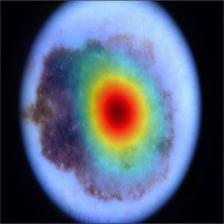} &\cincludegraphics[width=.08\textwidth]{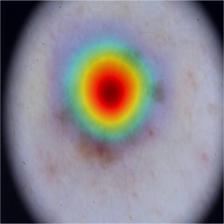}&\cincludegraphics[width=.08\textwidth]{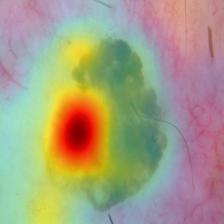}&\cincludegraphics[width=.08\textwidth]{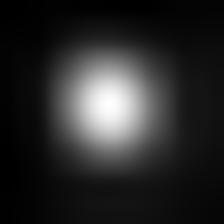}&\cincludegraphics[width=.08\textwidth]{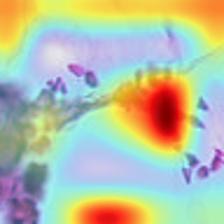}&\cincludegraphics[width=.08\textwidth]{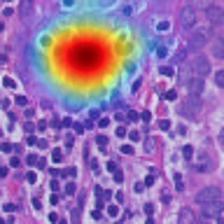}&\cincludegraphics[width=.08\textwidth]{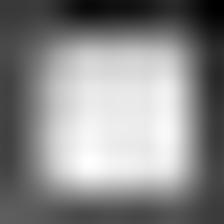}\\
        \midrule
        \multirow{5}{*}[-1.9cm]{\rotatebox[origin=c]{90}{AttentionMaps}} 
        & \multirow{2}{*}[-0.2cm]{\rotatebox[origin=c]{90}{ResNet}}
            & \rotatebox[origin=c]{90}{+G} & \cincludegraphics[width=.08\textwidth]{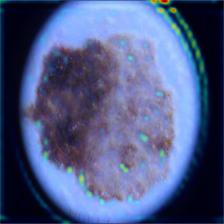} &\cincludegraphics[width=.08\textwidth]{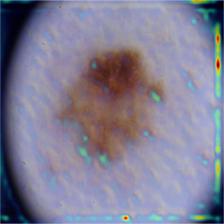}&\cincludegraphics[width=.08\textwidth]{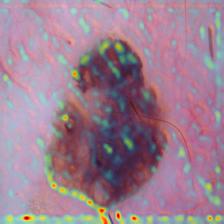}&\cincludegraphics[width=.08\textwidth]{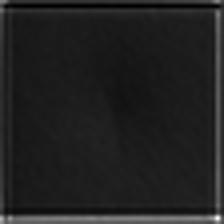}&\cincludegraphics[width=.08\textwidth]{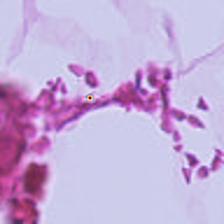}&\cincludegraphics[width=.08\textwidth]{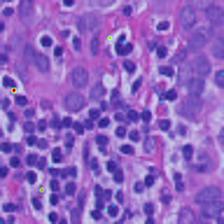}&\cincludegraphics[width=.08\textwidth]{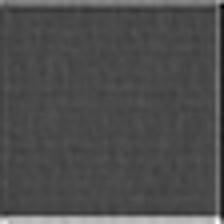}\\
            & & \rotatebox[origin=c]{90}{+GE} & \cincludegraphics[width=.08\textwidth]{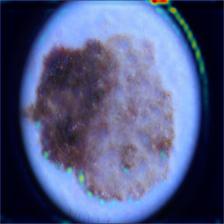} &\cincludegraphics[width=.08\textwidth]{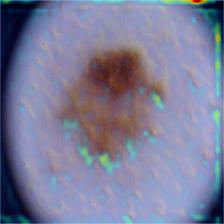}&\cincludegraphics[width=.08\textwidth]{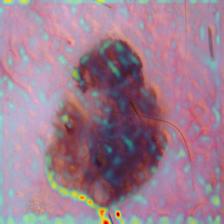}&\cincludegraphics[width=.08\textwidth]{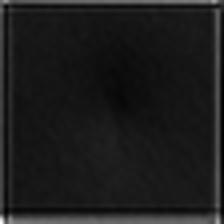}&\cincludegraphics[width=.08\textwidth]{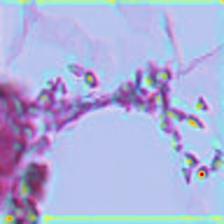}&\cincludegraphics[width=.08\textwidth]{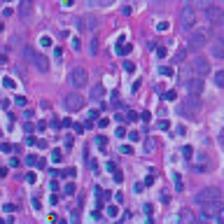}&\cincludegraphics[width=.08\textwidth]{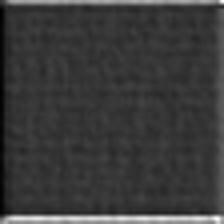}\\
            \cmidrule{2-10}
        & \multirow{2}{*}{\rotatebox[origin=c]{90}{EfficientNet}}
            & \rotatebox[origin=c]{90}{+G} & \cincludegraphics[width=.08\textwidth]{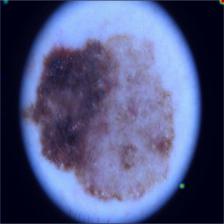} &\cincludegraphics[width=.08\textwidth]{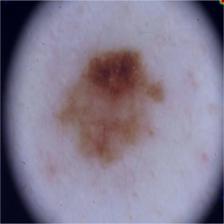}&\cincludegraphics[width=.08\textwidth]{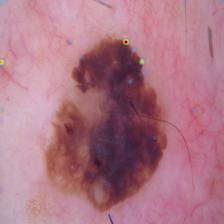}&\cincludegraphics[width=.08\textwidth]{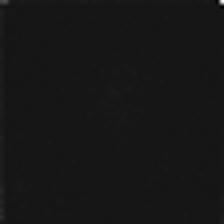}&\cincludegraphics[width=.08\textwidth]{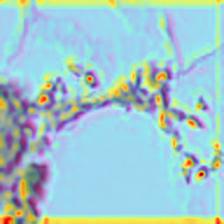}&\cincludegraphics[width=.08\textwidth]{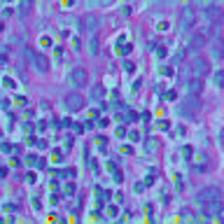}&\cincludegraphics[width=.08\textwidth]{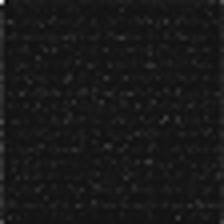}\\
            & & \rotatebox[origin=c]{90}{+GE} & \cincludegraphics[width=.08\textwidth]{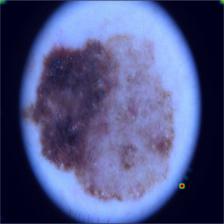} &\cincludegraphics[width=.08\textwidth]{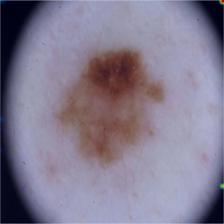}&\cincludegraphics[width=.08\textwidth]{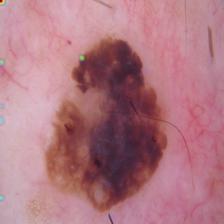}&\cincludegraphics[width=.08\textwidth]{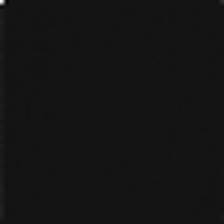}&\cincludegraphics[width=.08\textwidth]{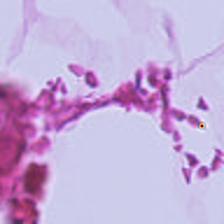}&\cincludegraphics[width=.08\textwidth]{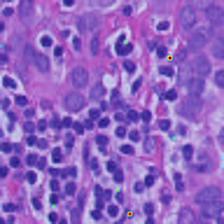}&\cincludegraphics[width=.08\textwidth]{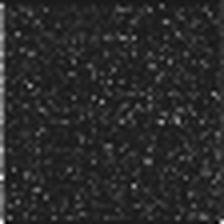}\\
            \cmidrule{2-10}
        \multicolumn{3}{c}{\rotatebox[origin=c]{90}{ViT}} & \cincludegraphics[width=.08\textwidth]{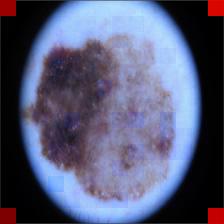} &\cincludegraphics[width=.08\textwidth]{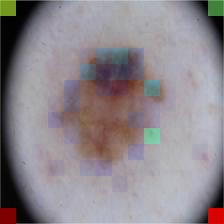}&\cincludegraphics[width=.08\textwidth]{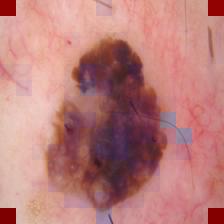}&\cincludegraphics[width=.08\textwidth]{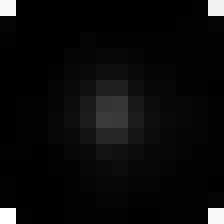}&\cincludegraphics[width=.08\textwidth]{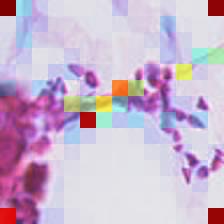}&\cincludegraphics[width=.08\textwidth]{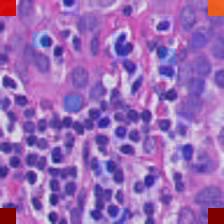}&\cincludegraphics[width=.08\textwidth]{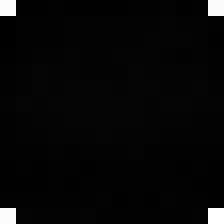}\\

    \end{tabular}
\end{table*}

\subsection{\uppercase{Local Explanations - Visualizations}}

Introducing self-attention into CNNs allows for a direct visualization of important image areas, which can help to gain further insights about the model decision process. 
We leverage the method from~\cite{Wang18:NLNN}, which creates a map showing how a certain pixel pays attention to all others and calculates the average over all of these maps.
We compare this attention map with Grad-CAM \cite{selvaraju2017grad} and the visualization method of the default ViT~\cite{kolesnikov:vit}.
Table~\ref{tab:att-vis} contains these visualizations generated using five images (all classes for both datasets) and additionally displays the mean of the attention maps over 500 images.
To get visually comparable results, we normalized the values between 0 and 255.
To stay comparable and only generate one visualization per image, we select the Grad-CAM for the predicted class for all examples.
We find that while models focus on different parts of the images, we observe no visible improvements or biases. 
Noteworthy, the highlighted area of Grad-CAM is typically larger than the one for the GA visualization due to the way Grad-CAM is implemented in \cite{selvaraju2017grad}, which we directly follow. Grad-CAM highlights mostly the whole object of interest. This can be seen in the mean image for the ISIC task because the center, where most of the skin lesions are placed, gets highlighted the most.
On the other hand, the attention map visualization highlights more specific features like the border of the skin lesion, framing both interpretability techniques as uniquely different. 
Interestingly GA, as well as the ViT, put a lot of attention on the corner of the image. While this is initially quite unintuitive, we suggest that this is similar behavior as reported in  \cite{darcet2023vision,xiao2023efficient}. Here, transformers often produce attention spikes in the image background, as visible in our mean images, to store information for further calculations. Unfortunately, this behavior cannot be further unraveled by looking at the attention maps, therefore, revealing a possible disadvantage of these visualizations in comparison to Grad-CAM.

Since we do not carry extensive medical knowledge, we refrain from more interpretations into the visualization here. 
We, however, note that the attention maps do not provide extensive visual explanations that could not be provided by Grad-CAM approaches in a similar fashion. 
Additionally, we do not observe specific structural biases since models seem to set a reasonable focus concerning the input data typically. 
Therefore, we attribute no systematic benefit to attention maps neither concerning local feature usage nor interpretability.

\section{\uppercase{Conclusions}}

This work investigated the usage of self-attention mechanisms for medical imaging, specifically classification. 
Toward this goal, we extended two widely used convolutional architectures with self-attention mechanisms and empirically compared these extended models against fully convolutional and attention-only baselines. 
We conducted this comparison on a dataset constructed from the well-known ISIC archive \cite{web:isic} and on the Camelyon17 dataset \cite{bandi2018:detection}.
Additionally, we analyze the OOD generalization on domain-shifted test sets.

While we did observe some minor improvements, none of the attention-featuring architectures led to a statistically significant increase in performance.
Sometimes, we even observe a significant decrease in balanced accuracy.
In some instances, we find the reduction of performance on the OOD test data is decreased when using self-attention.
However, we observe that this behavior is somewhat sensitive to the used backbone model, dataset, and in-distribution performance.
Our investigation suggests that self-attention alone does not result in a direct increase in performance or generalization of medical image classification models.

To explain our findings further, we conducted two related analyses, taking a closer look at explanations and feature usage of our architectures.
First, we performed a feature analysis of the best-performing skin lesion classification models using the attribution method described in \cite{reimers2020determining}.
We observed no systematic improvements in medically relevant features or bias features over the fully convolutional baselines, only slight deviations. 
Important global features, such as dermoscopic structures, are still not learned by employing self-attention.

Second, we performed a qualitative analysis of the learned attention maps. 
While including attention increases interpretability, e.g., revealing biased feature usage, we find that out-of-the-box explanation methods, we used Grad-CAM \cite{selvaraju2017grad}, perform similarly for that purpose. 
To conclude: our work indicates that merely including self-attention does not directly lead to benefits.
Of course, more work is required to ultimately conclude the suitability of self-attention mechanisms for medical image analysis. 
In any case, we hope to inspire future work to investigate new architectural changes more thoroughly than merely comparing performance scores since we believe this can provide valuable insights.

\bibliographystyle{apalike}
{\small
\bibliography{main}}

\section*{\uppercase{Appendix}}
Table~\ref{tab:hypers} contains the best hyperparameters, we determined for our models.
\begin{table}
    \centering
    \caption{Best determined hyperparameters}
    \begin{tabular}{ll|cc|cc}
        & & \multicolumn{2}{c|}{ISIC} & \multicolumn{2}{c}{Camelyon17}\\
        \multicolumn{2}{c|}{Model} & LR & WD & LR & WD \\
        \hline
        \multirow{6}{*}{\rotatebox[origin=c]{90}{ResNet}}  & Base & $0.001$ & $0.0001$ & $0.001$ & $0.001$\\
         & +GA & $0.001$ & $0$ & $0.001$ & $0.001$\\
         & +LA & $0.001$ & $0.0001$ & $0.0001$ & $0.01$\\
         & +GA+LA & $0.001$ & $0$ & $0.001$ & $0.0001$\\
         & +ELA & $0.001$ & $0$ & $0.01$ & $0$\\
         & +GA+ELA & $0.001$ & $0$ & $0.001$ & $0.0001$\\
        \hline
        \multirow{6}{*}{\rotatebox[origin=c]{90}{EfficientNet}}  & Base & $0.01$ & $0$ & $0.001$ & $0.0001$\\
         & +GA &  $0.01$ & $0$ & $0.001$ & $0.0001$\\
         & +LA &  $0.01$ & $0$ & $0.001$ & $0$\\
         & +GA+LA &  $0.01$ & $0$ & $0.001$ & $0.0001$\\
         & +ELA & $0.001$ & $0$ & $0.001$ & $0.001$\\
         & +GA+ELA &  $0.01$ & $0$ & $0.01$ & $0$\\
         \hline
         \multirow{2}{*}{\rotatebox[origin=c]{90}{ViT}} & 1K & $0.001$ & $0$ & $0.001$ & $0$\\
         & 21K & $0.0001$ & $0.0001$ & $0.001$ & $0.0001$\\
    \end{tabular}
    
    \label{tab:hypers}
\end{table}

\end{document}